\documentclass[10pt, a4paper,conference, final]{IEEEtran}
\usepackage{cite}
\usepackage{bm}
\usepackage{amsmath}
\usepackage{multirow}
\usepackage[pdftex]{graphicx}

\ifCLASSINFOpdf

\else

\fi

\usepackage{amssymb}
\usepackage{amsmath}
\usepackage{hhline}
\usepackage{multirow}

\usepackage{hyperref}

\begin{document}
%

\title{Free-Form Image Inpainting via Contrastive Attention Network}

\author{\IEEEauthorblockN{Xin Ma$^{1,2,3}$, Xiaoqiang Zhou$^{2,3,4}$, Huaibo Huang$^{1,3}$, Zhenhua Chai$^{2}$, Xiaolin Wei$^{2}$, Ran He$^*{^{1,3}}$}
\IEEEauthorblockA{{$^1$}School of Artificial Intelligence, University of Chinese Academy of Sciences\\
{$^2$}Vision Intelligence Center, AI Platform, Meituandianping Group\\
$^{3}$NLPR$\;\&\;$CEBSIT, CASIA \quad
{$^4$}University of Science and Technology of China\\
{\tt\small \{xin.ma, huaibo.huang\}@cripac.ia.ac.cn, xq525@mail.ustc.edu.cn}\\
{\tt\small \{chaizhenhua, weixiaolin02\}@meituan.com, rhe@nlpr.ia.ac.cn}}
} 

\maketitle
\renewcommand{\thefootnote}{\fnsymbol{footnote}}
\footnotetext{* indicates the correspondence author}


\begin{abstract}
Most deep learning based image inpainting approaches adopt autoencoder or its variants to fill missing regions in images. Encoders are usually utilized to learn powerful representational spaces, which are important for dealing with sophisticated learning tasks. Specifically, in image inpainting tasks, masks with any shapes can appear anywhere in images (i.e., free-form masks) which form complex patterns. It is difficult for encoders to capture such powerful representations under this complex situation. To tackle this problem, we propose a self-supervised Siamese inference network to improve the robustness and generalization. It can encode contextual semantics from full resolution images and obtain more discriminative representations. we further propose a multi-scale decoder with a novel dual attention fusion module (DAF), which can combine both the restored and known regions in a smooth way. This multi-scale architecture is beneficial for decoding discriminative representations learned by encoders into images layer by layer.  In this way, unknown regions will be filled naturally from outside to inside. Qualitative and quantitative experiments on multiple datasets, including facial and natural datasets (i.e., Celeb-HQ, Pairs Street View, Places2 and ImageNet), demonstrate that our proposed method outperforms state-of-the-art methods in generating high-quality inpainting results.
\end{abstract}

\IEEEpeerreviewmaketitle

\section{Introduction}
Image inpainting (a.k.a image completion or image hole-filling) aims at filling missing regions of an image with plausible contents \cite{bertalmio2000image}. It is a fundamental low-level vision task and can be applied to many real-world applications such as photo editing, distracting object removal, image-based rendering, etc \cite{yu2019free,barnes2009patchmatch,simakov2008summarizing}. The core goal of image inpainting is producing semantically meaningful contents in unknown areas, which can incorporate smoothly with known areas.

Traditional exemplar-based methods \cite{efros2001image,barnes2009patchmatch,xu2010image} (e.g., PatchMatch) can gradually synthesize plausible stationary contents by copying and pasting similar patches from known areas. The performances of them are satisfying when dealing with background inpainting tasks. But non-repetitive and complicated scenes, such as faces and objects are the Waterloo of these traditional methods because of limited ability of capturing high-level semantics. 

\begin{figure}[ht]
    \centering
    \includegraphics[scale=0.58]{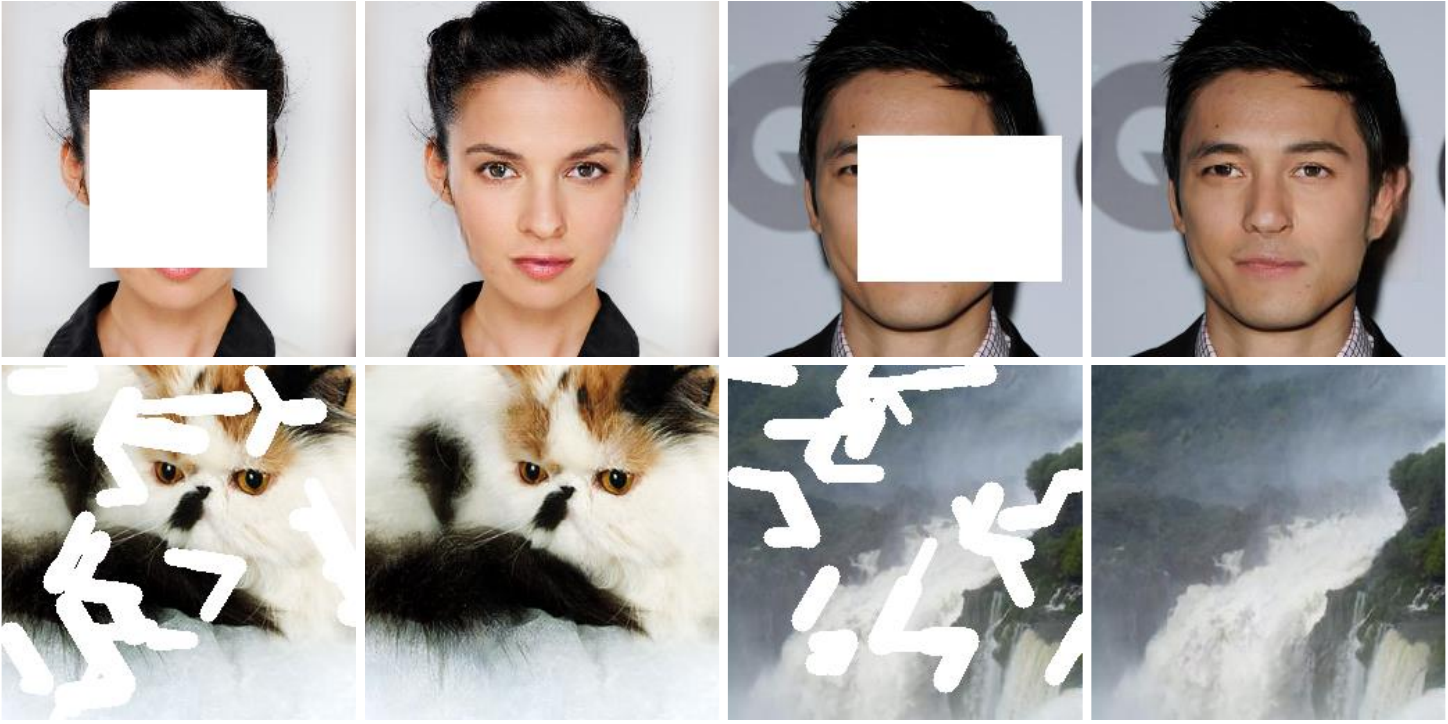}
    \caption{Images with free-form masks including rectangular and irregular masks and corresponding inpaining results by using our proposed method.}
    \label{first fig}
\end{figure}

Recently, deep convolutional neural networks (CNNs) have made great progress in many computer vision tasks \cite{finn2017model,liu2018darts, cao2018learning, huang2019wavelet}. Benefiting from the powerful ability of representation learning of CNNs, many deep learning based methods have been proposed. These approaches adopt autoencoder or its variants 
architecture jointly trained with generative adversarial networks (GANs) to hallucinate semantically plausible contents in unknown regions \cite{yu2019free,xie2019image,liu2018image}.

Specifically, masks can be of intricate and irregular patterns and can appear anywhere in images, which greatly increase the difficulty of image inpainting. Previous image inpainting approaches jointly train an encoder and a decoder by some common loss functions (e.g., reconstruct loss, style loss, etc). It is difficult for the encoder to learn powerful representational spaces form images with free-form masks. As a result, these CNN-based approaches will produce depressing results with obvious color contrasts and artifacts especially in boundary areas. A naive way is to design a very deep network to obtain a larger model capacity. However, it heavily increases the computational cost and may not learn accurate latent representational spaces.

For handling this limitation, we propose a self-supervised Siamese inference network with contrastive learning. As explored in \cite{he2020momentum, hadsell2006dimensionality}, contrastive learning trains an encoder to perform a dictionary $look-up$ task. A 'query' encoded by one of the encoders should be similar with its corresponding 'key' (token) that is sampled from data (e.g., image or patch) and is usually represented by another encoder. We assume that two identical images with different masks consist a positive pair while two different images form a negative pair. In order to acquire large and consistent dictionaries that are benefit for representation learning, we follow MoCo \cite{he2020momentum} to design a queue dictionary and a momentum-update based key encoder. In this way, the robustness and the accuracy (i.e., not producing a noisy representation of the input image) of the encoder can be improved.

Many previous approaches consider image inpainting as a conditional image generation task. The roles of the encoder and decoder are recognizing high-level semantic information and synthesizing low-level pixels \cite{yu2018generative}, respectively. These approaches, e.g., PConv \cite{liu2018image}, LBAM \cite{xie2019image} and Yu's method \cite{yu2019free}, focus more on missing areas and synthesize realistic alternative contents by a well-designed attention architecture or some effective loss functions. However, due to less attention to structural consistency, there are obvious color contrasts, or artificial edge responses, especially in boundaries of results produced by these methods. In fact, the progress of biology inspired us that the human visual system is sensitive to disharmonious transition regions. Therefore, we pay more attention to the structural continuity of restored images surrounding holes while generating texture-rich images. 


To properly suppress color discrepancy and artifacts in boundaries, we propose a novel and independent dual attention fusion module (DAF) to synthesize pixel-wise smooth contents, which can be inserted into autoencoder architectures in a plug-and-play way. The core idea of the fusion module is to calculate the similarity between the synthesized content and the known region. Some methods are proposed to address this problem such as DFNet \cite{hong2019dfnet} and Perez's method \cite{perez2003poisson}. But DFNet lacks flexibility in handing different information types (e.g., different semantics), hindering more discriminative representations. Our proposed DAF is developed to adaptively recalibrate channel-wise feature by taking interdependencies between channels into account while force CNNs focusing more on unknown regions. DAF will provide a combining map to blend restored contents and original images in a smooth way.

Qualitative and quantitative experiments on multiple datasets including facial and natural datasets (i.e., Celeb-HQ, Paris SteetView, Places2 and ImageNet) are utilized to evaluate our proposed method. The experimental results demonstrate that our proposed method outperforms state-of-the-art methods in generating high-quality inpainting results. To sum up, the main contributions of this paper are as follows:
\begin{itemize}
    \item We propose a Siamese inference network based on contrastive learning for free-form image inpainting. It helps to improve the robustness and accuracy of representation learning for complex mask patterns.
    \item We propose a novel independent dual attention fusion module that can explore feature interdependencies in spatial and channel dimensions and provide a combine map. Smooth contents with rich texture information can be naturally synthesized from outside to inside.
    \item Our proposed method achieves smooth inpainting results with richer texture information on four standard datasets against state-of-the-art image inpainting methods.
\end{itemize}

\section{Related Work}
\subsection{Image Inpainting}
Traditional image inpainting methods are mainly diffusion-based \cite{bertalmio2000image} or patch-based \cite{barnes2009patchmatch}. Bertalmio \textit{et al}. \cite{bertalmio2000image} proposed a algorithm to propagate appearance information from the neighboring region to the unknown regions. In PatchMatch \cite{barnes2009patchmatch}, a fast nearest neighbor searching algorithm is proposed to search and paste the most similar image patches from the 
known regions. These methods utilize low-level image features to guide the feature propagation from known image background or image datasets to corrupted regions. These methods work well when holes are small and narrow, or there are plausible matching patches in known region. However, when suffering from complicated scenes, it is difficult for these approaches to produce semantically plausible solutions, due to a lack of semantic understanding of images.

To accurately recover damaged images, many methods \cite{pathak2016contextencoder, zhang2017demeshnet, yu2018contextualattention, gmcnn, gatedconviccv} adopt deep convolutional neural networks (CNNs) \cite{wang2018pix2pixHD}, especially generative adversarial networks (GANs) \cite{goodfellow2014gan} in image inpainting. Context Encoder \cite{pathak2016contextencoder} formulates image inpainting as a conditional image generation problem. Global-Local \cite{iizuka2017globalandlocal}
 utilizes two discriminators to improve the quality of the generated images at different scales, facilitating both globally and locally consistent image completion. Some works \cite{yu2018contextualattention, nazeri2019edgeconnect, ren2019structureflow} design a coarse-to-fine framework to solve sub-problem of image inpainting in different stages, e.g., Edge connect\cite{nazeri2019edgeconnect} firstly recovers the edge map of the corrupted image and generates image textures in the second stage.
 
\subsection{Self-Supervised Representation Learning}
Self-supervised representation learning has shown great potential recently in many research works \cite{he2020momentum, zhan2020self, chen2020SimCLR}. Compared with supervised learning, self-supervised learning utilizes unlabeled data to learn representations. In MoCo \cite{he2020momentum} and SimCLR \cite{chen2020SimCLR}, good representations are learned by contrasting positive pairs against negative pairs. In MoCo \cite{he2020momentum}, the authors use a self-supervised learning strategy and achieve comparable performance with supervised methods. Self-supervised learning strategy is also used in many other vision tasks recently. Mustikovela \textit{et al.} \cite{mustikovela2020self-viewpoint} used the self-supervised learning to viewpoint learning by taking use of generative consistency and symmetry constraint. Zhan \textit{et al.} \cite{zhan2020self-occlusion} utilized a mask completion network to predict occlusion ordering, which is trained with self-supervised learning strategy.

\begin{figure*}[ht]
    \centering
    \includegraphics[scale=0.75]{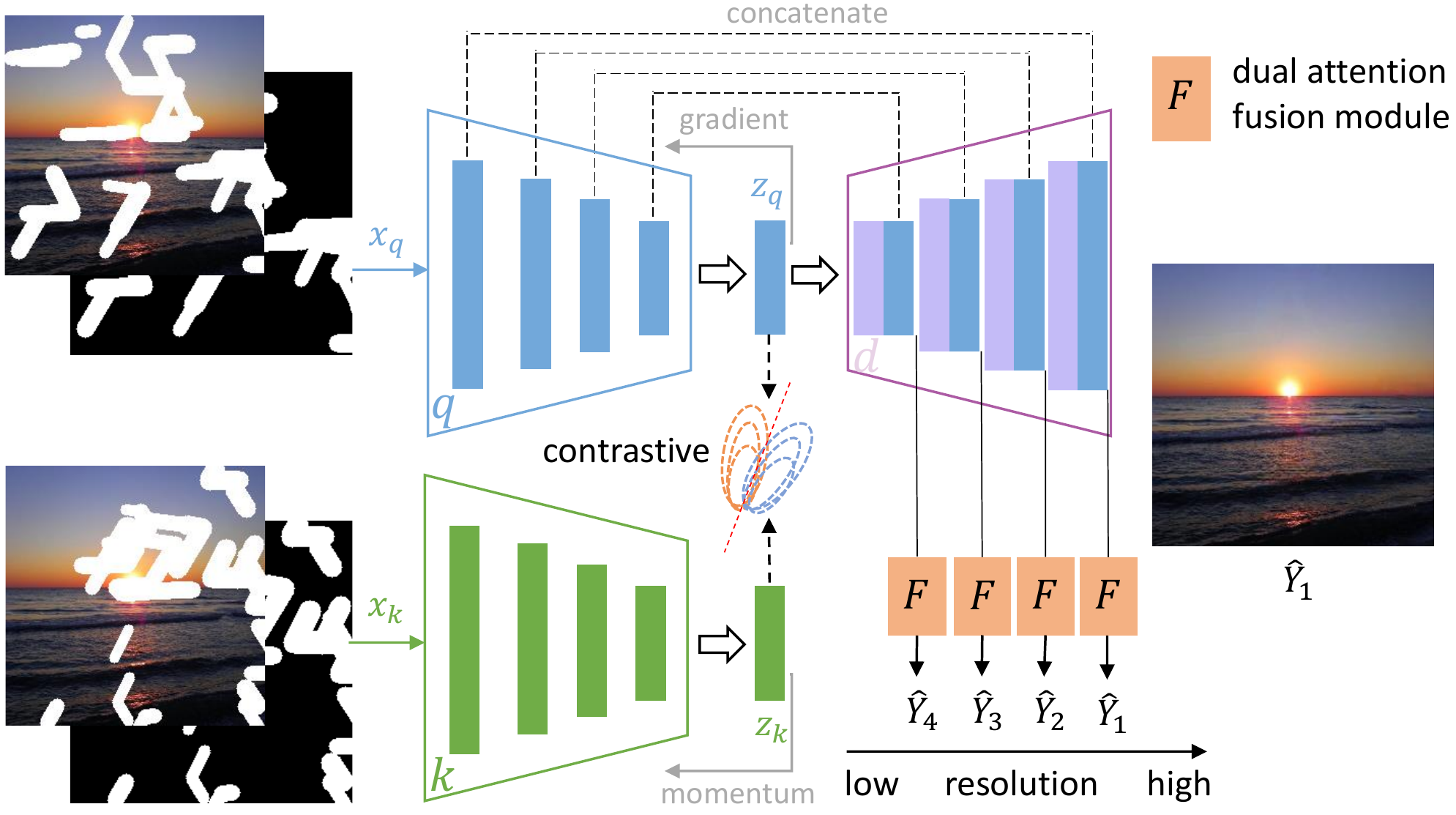}
    \caption{The network architecture of our method. The self-supervised Siamese inference network consists of encoder $E_q$ and $E_k$ with contrastive learning. This inference network encodes the new key representations on-the-fly by using the momentum-updated encoder $E_k$. We insert the dual attention fusion module into several decoder layers, thus forming a multi-scale decoder. The inference network is firstly trained on ImageNet with contrastive learning. Then the pre-trained encoder $E_q$ and the decoder are jointly trained with the fusion module.}
    \label{pipeline architecture}
\end{figure*}

\subsection{Attention Mechanism}
Attention mechanism is a hot topic in computer vision and has been widely investigated in many works \cite{wang2018non-local,dai2019second, fu2019dualattention, hu2018senet}. The ways to utilize attention mechanism can be coarsely divided into two categories: spatial attention \cite{wang2018non-local} and channel attention \cite{hu2018senet}. Yu \textit{et al.} \cite{yu2018contextualattention} proposed a contextual attention to calculate the spatial attention scores between pixels in corrupted region and known region. DFNet \cite{hong2019dfnet} utilizes a spatial alpha composition map to combine features in the corrupted region and known region. In this paper, we investigate both spatial attention and channel attention mechanism to further improve the performance of image inpainting.

\section{Methodology}
In this section, we first present our self-supervised Siamese inference network. Subsequently, the details of the dual attention fusion (DAF) module and learning objectives in our method are provided. The overall framework of our image inpainting method is shown in Fig. \ref{pipeline architecture}.

\subsection{Self-supervised Siamese inference network}
Our proposed self-supervised Siamese inference network consists of two identical encoders but not sharing parameter weights \cite{hadsell2006dimensionality,wang2015unsupervised,he2020momentum}, noted as $E_q$ and $E_k$, respectively. The proposed inference network is trained by contrastive learning, which can be seen as training an encoder to perform a dictionary look-up task: a 'query' encoded by $E_q$ should be similar with its corresponding 'key' (i.e., positive key) represented by another encoder $E_k$ and dissimilar to others (i.e., negative keys). Two identical images with different masks are required for the proposed inference network, named as $x_q$ and $x_k$, respectively. Thus, we can obtain a query representation $z_q=E_q(x_q)$ and a key representation $z_k=E_k(x_k)$, respectively. Followed many previous self-supervised works \cite{bachman2019learning,zhuang2019local}, the contrastive loss is utilized as self-supervised objective function for training the proposed inference network and can be written as:
\begin{equation}
    \mathcal{L}=-log\frac{exp(z_q.z_k^+/\tau)}{\sum_{i=0}^{k}exp(z_q.z_{k_i}/\tau)},
\end{equation}
where $\tau$ is the temperature hyper-parameter and it degrades into the original $softmax$ when $\tau$ is equal to 1. The output will be less sparse with $\tau$ increasing \cite{chen2020dynamic}. The $\tau$ is set as $0.07$ for efficient training process in this work. Specially, this loss, also known as InfoNCE loss\cite{hadsell2006dimensionality,he2020momentum}, tries to classify $z_q$ as $z_k^+$.

High-dimensional continuous images can be projected into a discrete dictionary by contrastive learning. There are three general mechanisms for implementing contrastive learning (i.e., end-to-end \cite{hadsell2006dimensionality}, memory bank \cite{wu2018unsupervised} and momentum updating \cite{he2020momentum}), whose main differences are how to maintain keys and how to update the key encoder. Considering GPU memory size and powerful feature learning, we follow MoCo \cite{he2020momentum} to design a consistent dictionary implemented by $queue$. Thus, the key representations of the current batch data are enqueued into the dictionary while the older representations are 
dequeued progressively. The length of the queue is under control, which enable the dictionary to contain a large number of negative image pairs. Specially, representation learning can be beneficial from a dictionary with a large scale negative pairs. We set the length of the queue as $65536$ in this work.

It is worth noting that the encoder $E_k$ is updated by a momentum update strategy instead of direct back-propagation. The main reason is that it's difficult to propagate the gradient to all keys in the queue. The updating process of $E_k$ can be formulated as follows:
\begin{equation}
    \theta_k \leftarrow m\theta_k+(1-m)\theta_q,
\end{equation}
where $\theta_q$ and $\theta_k$ are denoted as the parameters of $E_q$ and $E_k$, respectively. $\theta_q$ is updated by back-propagation. $m\in[0,1)$ is the momentum coefficient hyper-parameter and set as $0.9$ in this paper. In this way, the encoder $E_k$ can update smoothly relative to $E_q$, resulting in a more consistent discrete dictionary.

\subsection{Dual Attention Fusion Module}
\begin{figure}[ht]
    \centering
    \includegraphics[scale=0.58]{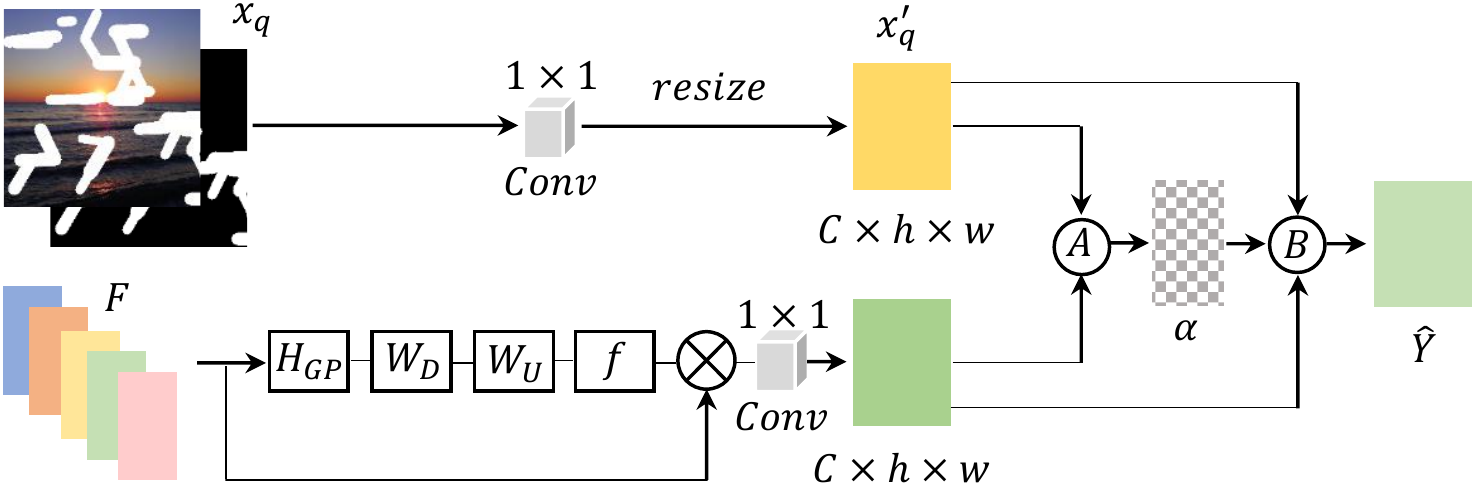}
    \caption{The network architecture of the dual attention fusion module. It first predicts an $\alpha$ combine map with the learnable transformation function $\mathcal{A}$. Then we can obtain final smooth inpainting results with rich texture by the combine function $\mathcal{B}$.}
    \label{fusion}
\end{figure}

We now give more details about our proposed dual attention fusion module (see Fig. \ref{fusion}), which contains a channel attention mechanism and a spatial attention mechanism. This fusion module are embedded to the last few layers of the decoder and output inpainting results with multi-scale resolutions \cite{karras2017progressive,brock2018large}. Thus, constraints can be imposed on multi-scale outputs for high-quality results.

Previous CNN-based image inpainting approaches treat channel-wise features equally, thus hindering the ability of the representation learning of the network. Meanwhile, high-level and interrelated channel features can be considered as specific class responses. For more discriminative representations, we first build a channel attention module in our proposed fusion module. 

As shown in Fig. \ref{fusion}, let a feature map $F=[f_1,\cdots,f_c,\cdots,f_C]$ be one of the inputs of the fusion module, whose channel number is $c$ and size is $h \times w$. The channel descriptor can be acquired from the channel-wise global spatial information by global averaging pooling. Then we can obtain the channel-wise statistics $z_c \in \mathbb{R}^c$ by shrinking $F$:
\begin{equation}
    z_c=H_{GP}(x_c)=\frac{1}{h \times w}\sum_{i=1}^{h}\sum_{j=1}^{w}f_c(i,j),
\end{equation}
where $z_c(i,j)$ is the $c$-th element of $z$. $H_{GP}$ means the global pooling function.

In order to fully explore channel-wise dependencies from the aggregated information by using global average pooling, we introduce a gating mechanism. As illustrated in \cite{hu2018senet,zhang2018image}, the sigmoid function can be used as a gating function:
\begin{equation}
    \omega=f(W_U\delta(W_Dz)),
\end{equation}
where $f(\cdot)$ and $\delta(\cdot)$ are the sigmoid gating and ReLU functions, respectively. $W_D$ and $W_U$ are the weight sets of $Conv$ layers. They set channel number as $C/r$ and $C$, respectively. Finally, the channel statistics $\omega$ is acquired and used to rescale the input $f_c$:
\begin{equation}
    \hat{f_c} = w_c\cdot f_c,
\end{equation}
where $w_c$ and $f_c$ are the scaling factor and feature map of the $c$-th channel, respectively.

The long-range contextual information is essential for discriminant feature representations. We propose a spatial attention module that forms the final part of the proposed fusion module. Given an input image with a mask $x_q$, we first get $x{_q}^{'}$ that matches the size of the re-scaled feature map $\hat{F} \in \mathbb{R}^{c \times h \times w}$,
\begin{equation}
    x{_q}^{'}=(W_Cx_q)\downarrow,
\end{equation}
where $W_C$ and $\downarrow$ are the weight set of a $1 \times 1$ convolutional layer and down-scaled module, respectively. 

Then the combining map $\alpha \in \mathbb{R}^{C \times h \times w}$ is given by,
\begin{equation}
    \alpha=f(\mathcal{A}(W_D\hat{F},x{_q}^{'})),
\end{equation}
where $W_D$ is the weight set of a $1 \times 1$ convolutional layer. It sets channel number of $\hat{F}$ to be same with $x{_q}^{'}$. $\mathcal{A}$ is a learnable transformation function implemented by three $3 \times 3$ convolutional layers. $W_D\hat{F}$ and $x{_q}^{'}$ are first concatenated and then fed into convolutional layers. $f(\cdot)$ is the sigmoid function that can make $\alpha$ an attention map to some extent.

The final inpainting result $\hat{Y}$ is obtained by, 
\begin{equation}
    \hat{Y}=\mathcal{B}(\alpha,W_D\hat{F},x{_q}^{'})=\alpha\odot{W}_D \hat{F}+(1-\alpha)\odot{x}{_q}^{'},
\end{equation}
where $\odot$ and $\mathcal{B}$ denote the Hadamard product and combine function, respectively. In this way, we can eliminate obvious color contrasts and artifacts especially in boundary areas and get a smoother inpainting results with richer texture.

\subsection{Loss Functions}
As explored in \cite{xie2019image,gatedconviccv,hong2019dfnet}, for synthesizing richer texture details and correct semantics, the element-wise reconstruction loss, the perceptual loss \cite{johnson2016perceptual}, the style loss, the total variation loss and the adversarial loss are used to train our proposed method. 

\textbf{Reconstruction Loss.} It is calculated as $\mathcal{L}_1$-norm between the inpainting result $\hat{Y}$ and the target image $Y$,
\begin{equation}
    \mathcal{L}_{rec}=||Y-\hat{Y}||_1.
\end{equation}

\textbf{Perceptual Loss.} Due to lacking high-level semantics by only using the element-wise loss, we introduce the perceptual loss,
\begin{equation}
    \mathcal{L}_{per}=\frac{1}{N}\sum^N_{i=1}||\Phi^i(Y)-\Phi^i(\hat{Y})||_1,
\end{equation}
where $\Phi$ is the VGG-16 network that pre-trained on ImageNet \cite{deng2009imagenet}. $\Phi^i(\cdot)$ outputs feature maps of the $i$-th pooling layer. We select $pool-1$, $pool-2$ and $pool-3$ layers of the pre-trained VGG-16 in this work.

\begin{figure*}[ht]
    \centering
    \includegraphics[scale=0.7]{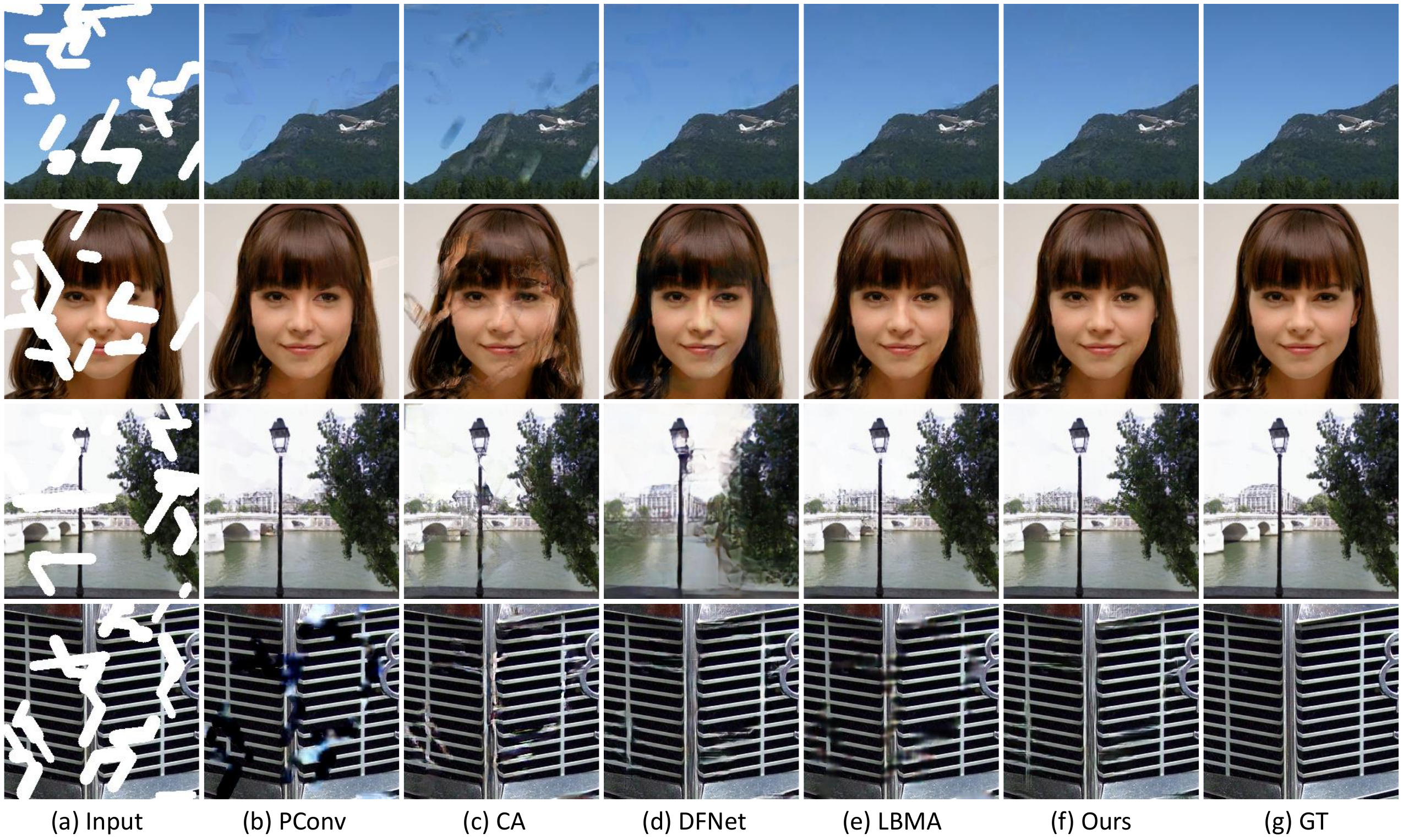}
    \caption{Qualitative experiments compared with state-of-the-arts on four datasets with free-form masks. (a) is the input with irregular mask. (b), (c), (d), (e) and (f) are the results generated by PConv \cite{liu2018image}, CA \cite{yu2018generative}, DFNet \cite{hong2019dfnet}, LBMA \cite{xie2019image} and ours method respectively from left to right. (g) is the ground truth.}
    \label{comaparison experments}
\end{figure*}

\textbf{Style Loss.} For getting richer textures, we also adopt the style loss defined on the feature maps produced by the pre-trained VGG-16. Followed \cite{xie2019image,liu2018image}, the style loss can be calculated as $\mathcal{L}_1$-norm between the Gram matrices of the feature maps,
\begin{equation}
    \mathcal{L}_{style}=\frac{1}{N}\sum^N_{i=1}\frac{1}{C_i \cdot C_i}||\Phi^i(Y)(\Phi^i(Y))^T-\Phi^i(\hat{Y})(\Phi^i(\hat{Y}))^T||_1,
\end{equation}
where $C_i$ denotes the channel number of the feature map at $i$-th layer in the pre-trained VGG-16.

\textbf{Total Variation Loss.} It is a smoothing penalty term and calculated on the region of 1-pixel dilation of the unknown regions.
\begin{equation}
\begin{aligned}
    \mathcal{L}_{tv}&=\frac{1}{N}\sum_{(i,j),(i,j+1)\in\Omega}||\hat{Y}^{i,j+1}-\hat{Y}^{i,j}||_1\\&+\frac{1}{N}\sum_{(i,j),(i+1,j)\in\Omega}||\hat{Y}^{i+1,j}-\hat{Y}^{i,j}||_1,
\end{aligned}
\end{equation}
where $\Omega$ means the unknown regions.

\textbf{Adversarial Loss.} We introduce the adversarial loss for improving the visual quality of the generated inpainting image. Followed \cite{gulrajani2017improved}, it can be formulated as,
\begin{equation}
\begin{aligned}
    \mathcal{L}_{adv}&=\min\limits_G\max\limits_D\mathbb{E}_{Y \sim P_{Y}}D(Y)-\mathbb{E}_{\hat{Y} \sim P_{\hat{Y}}}D(\hat{Y})\\&+\lambda\mathbb{E}_{{Y}^{'} \sim P_{{Y}^{'}}}((||\nabla_{{Y}^{'}}D({Y}^{'}||)^2)-1)^2,
\end{aligned}
\end{equation}
where $D(\cdot)$ means the discriminator. ${Y}^{'}$ is the resized version that sampled from $\hat{Y}$ and $Y$ by interpolation with a random scale factor. We set $\lambda$ as 10 in this work.

\textbf{Model Objective.} Taking the above loss functions and the multi-scale network architecture into account, we group them into two categories: $Structure$ $Loss$ and $Texture$ $Loss$, respectively. 
\begin{equation}
    \mathcal{L}_{struct}^k=\lambda_{rec}\mathcal{L}_{rec}^k,
\end{equation}
where $\lambda_{rec}$ means the weight factor and set as $6$. $\mathcal{L}_{struct}^k$ is calculated as $\mathcal{L}_{rec}$ at the $k$-th layer of the decoder ($1_{st}$ means the last decoder layer). 

The $Texture$ $Loss$ is given by,
\begin{equation}
    \mathcal{L}_{text}^{k}=\lambda_{per}\mathcal{L}_{per}^{k}+\lambda_{style}\mathcal{L}_{style}^{k}+\lambda_{tv}\mathcal{L}_{tv}^{k}+\lambda_{adv}\mathcal{L}_{adv}^{k},
\end{equation}
where $\lambda_{per}$, $\lambda_{style}$, $\lambda_{tv}$, $\lambda_{adv}$ are trade-off factors and set as $0.1$, $240$, $0.1$ and $0.001$ respectively in this work.

Finally, the total model objective is given by,
\begin{equation}
    \mathcal{L}_{total}=\frac{1}{|P|}\sum_{k \in P}\mathcal{L}_{struct}^k+\frac{1}{|Q|}\sum_{k \in Q}\mathcal{L}_{text}^k,
\end{equation}
where both $P$ and $Q$ are the selected decoder layer sets that imposed constraints. We select $P$ as $\{1,2,3,4,5,6\}$ and $Q$ as $\{1,2,3\}$ respectively for better quality inpainting results. Note that $1$ represents the outermost layer.

\section{Experiments}
To demonstrate the superiority of our approach against state-of-the-art image inpainting methods, both quantitative and qualitative experiments are conducted. In this section, we will introduce the details of our experimental settings and the experimental results one by one.

\subsection{Experiments Settings}
\textbf{Datasets.} We conduct experiments on several public datasets including:
\begin{itemize}
\item[-] CelebA-HQ \cite{karras2017progressive}, a dataset that contains 30,000 high-quality face images.
\item[-] Places2 \cite{zhou2017places}, a dataset that contains over 8,000,000 images from over 365 scenes collected from the natural world.
\item[-] Paris Street View \cite{doersch2012makes}, a dataset which consists of 15,000 images collected from street views of Paris
\item[-] ImageNet \cite{deng2009imagenet}, a large visual database containing more than 14 million images of the world.
\end{itemize}

For Places2, Pairs Street View and ImageNet, we use their origin splits for validation and testing. As for CelebA-HQ, we randomly select 28000 images for training and the rest for testing.

\begin{figure}[ht]
    \centering
    \includegraphics[scale=0.75]{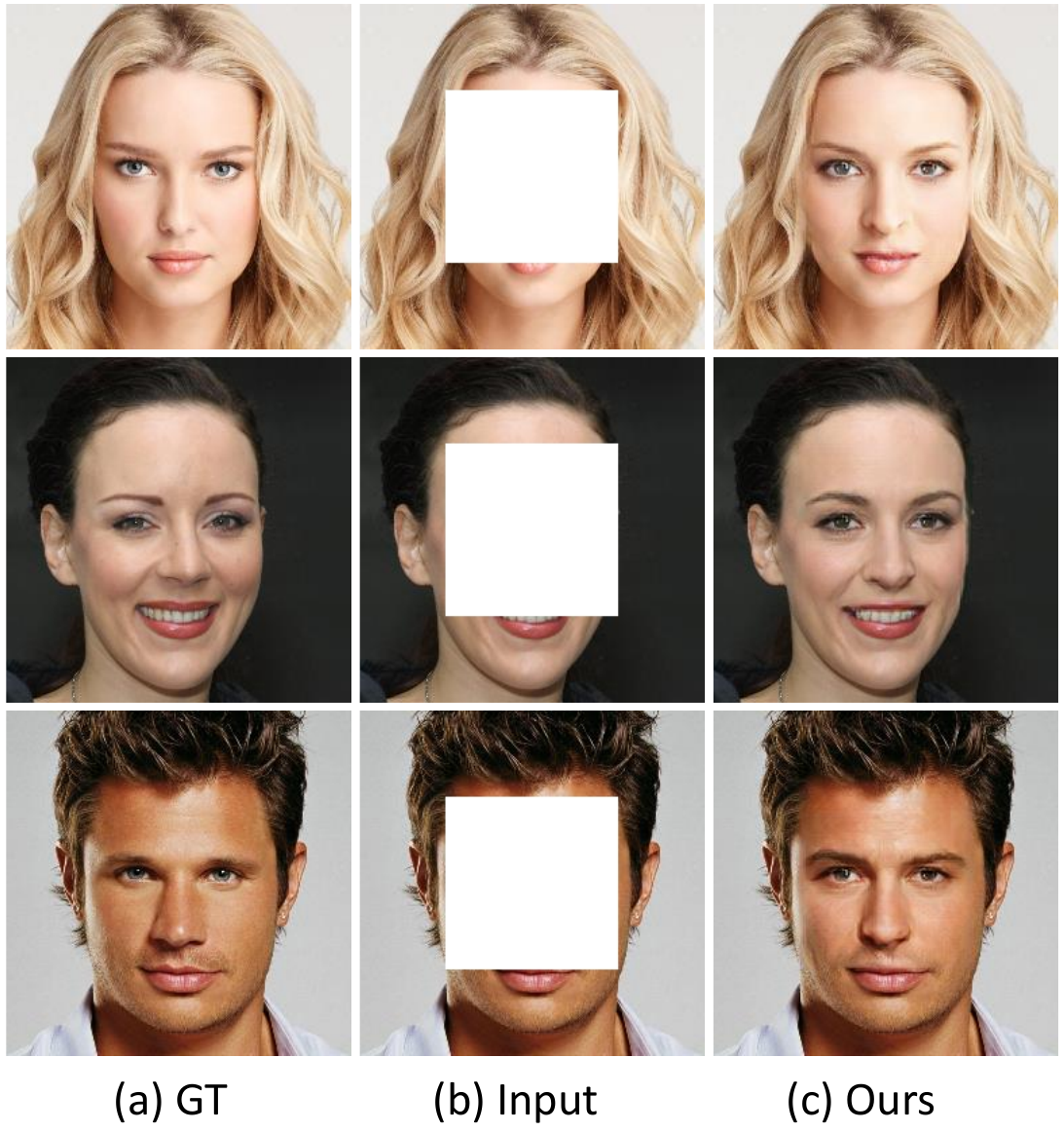}
    \caption{Example results predicted by our proposed method on CelebA-HQ.}
    \label{celba-hq}
\end{figure}

\textbf{Implementation Details.} All the images are resized to $256 \times 256$ during training and testing. Data augmentation is utilized during training such as filpping. Masks are free-form including rectangular and irregular and automatically generated on-the-fly during training \cite{gatedconviccv}. Taken as a whole, our proposed method can be broken down into two stages. In the first stage, the inference network is trained on ImageNet through contrastive learning until convergence. And in the next stage, the pre-trained encoder and the decoder are jointly trained with the fusion module. We use the SGD optimizer with the learning rate as 0.015 for training the Siamese inference network while use the Adam optimizer with the learning rate as $10^{-4}$ for jointly training the encoder and decoder. All the results are reported directly from the trained model without any additional post-processing.

\begin{table}[!t]
\begin{center}
\begin{tabular}{cccccc}
\hline
Method & Mask Type                    & $\mathcal{L}_1$ error\dag & PSNR\ddag                      & SSIM\ddag   & FID\dag                      \\ \hline\hline
PConv \cite{liu2018image} & \multirow{5}{*}{rectangular}  & 0.0357    & 22.83                     & 0.7737   & 26.45                    \\
CA \cite{yu2018generative}     &                              & 0.0372    & 22.58                     & 0.7800 & 21.78                    \\
DFNet \cite{hong2019dfnet} &                              & 0.0349    & 22.95                     & 0.7789 & 26.73                    \\
LBMA \cite{xie2019image} &                              & 0.0350    & 23.08                     & 0.7817   & 24.01                    \\
Ours   &                              & \textbf{0.0332}    & \textbf{23.52}                     & \textbf{0.7928} & \textbf{21.42}                    \\ \hline
PConv \cite{liu2018image} & \multirow{5}{*}{irregular}   & 0.0202    & 26.65                     & 0.9224 & 12.03                     \\
CA \cite{yu2018generative}    &                              & 0.0384    & 22.08                     & 0.8182  & 28.69                    \\
DFNet \cite{hong2019dfnet} &                              & 0.0180    & 27.40                     & 0.9343 & 10.12                     \\
LBMA \cite{xie2019image}  &                              & 0.0179    & 27.43 & 0.9351 & 10.56                    \\
Ours   &                              & \textbf{0.0178}    & \textbf{27.54} & \textbf{0.9350} & \textbf{9.15} \\ \hline
\end{tabular}
\end{center}
\caption{Quantitative comparison on validation images of Places2 with both rectangle and irregular masks. \dag Lower is better. \ddag Higher is better.}
\label{quantitative results}
\end{table}

\subsection{Quantitative Results}
We conduct quantitative experiments on the Places2 dataset with free-form masks. The $L_1$ $loss$, peak signal-to-noise ratio (PSNR), structural similarity index (SSIM) and Fre$^{'}$chet Inception Distance (FID) are used for evaluation metrics. 

The ability of the model to recover the original content in the hole can be roughly reflected by $\mathcal{L}_1$ $error$. PSNR and SSIM measure the similarity between the inpainting result and the target image. As for FID, it can measure the Wasserstein-2 distance between real and inpainting images through the pre-trained Inception-V3.

Table \ref{quantitative results} shows the performance of our proposed method against other state-of-the-art methods. Our method outperforms all the other compared methods in four metrics. The main reasons are that 1): images with free-form masks dramatically increase the difficulty of image inpainting, thus hindering the ability of the representation learning of the encoder; 2): exiting methods firstly take generating realistic images into account but ignore the structural consistency of the generated image. 

\subsection{Qualitative Results}
We compare our proposed method with state-of-the-art methods in term of visual and semantic coherence. We conduct qualitative experiments on the test set of four datasets with free-form masks. As shown in Fig. \ref{comaparison experments}, we mask the test images with irregular masks. It can be seen that PConv, CA, DFNet and LBMA tend to synthesise blurred and unsmooth final results. Our proposed method can generate smoother inpainting results with reasonable semantics and richer textures with the help of the self-supervised Siamese inference network and the DAF module. It demonstrates that our proposed method is superior to the comparison methods in terms of consistent structures and colors. Furthermore, as shown in Fig. \ref{celba-hq}, we also conduct experiments on the test images of CelebA-HQ with typical rectangular squares to evaluate the inpainting ability of our proposed method. Our method can generate face images with consistent colors and structures. 

\begin{table}[!t]
\begin{center}
\begin{tabular}{ccccc}
\hline
contrastive learning & $\times$ & $\checkmark$ & $\times$ & $\checkmark$ \\
DAF                  & $\times$ & $\times$ & $\checkmark$ & $\checkmark$ \\ \hline\hline
$\mathcal{L}_1$ error\dag              & 0.0192 & 0.0190 & 0.0186 & \textbf{0.0178} \\
PSNR\ddag              & 26.98 & 27.13 & 27.21 & \textbf{27.54} \\
SSIM\ddag                  & 0.9282 & 0.9302 & 0.9315 & \textbf{0.9351} \\
FID\dag                  & 12.62 & 12.56 & 11.14 & \textbf{9.15} \\ \hline
\end{tabular}
\end{center}
\caption{Ablation study experiments on on validation images of Places2. \dag Lower is better. \ddag Higher is better.}
\label{ablation study}
\end{table}

\subsection{Ablation Study}
\begin{figure}[ht]
    \centering
    \includegraphics[scale=0.38]{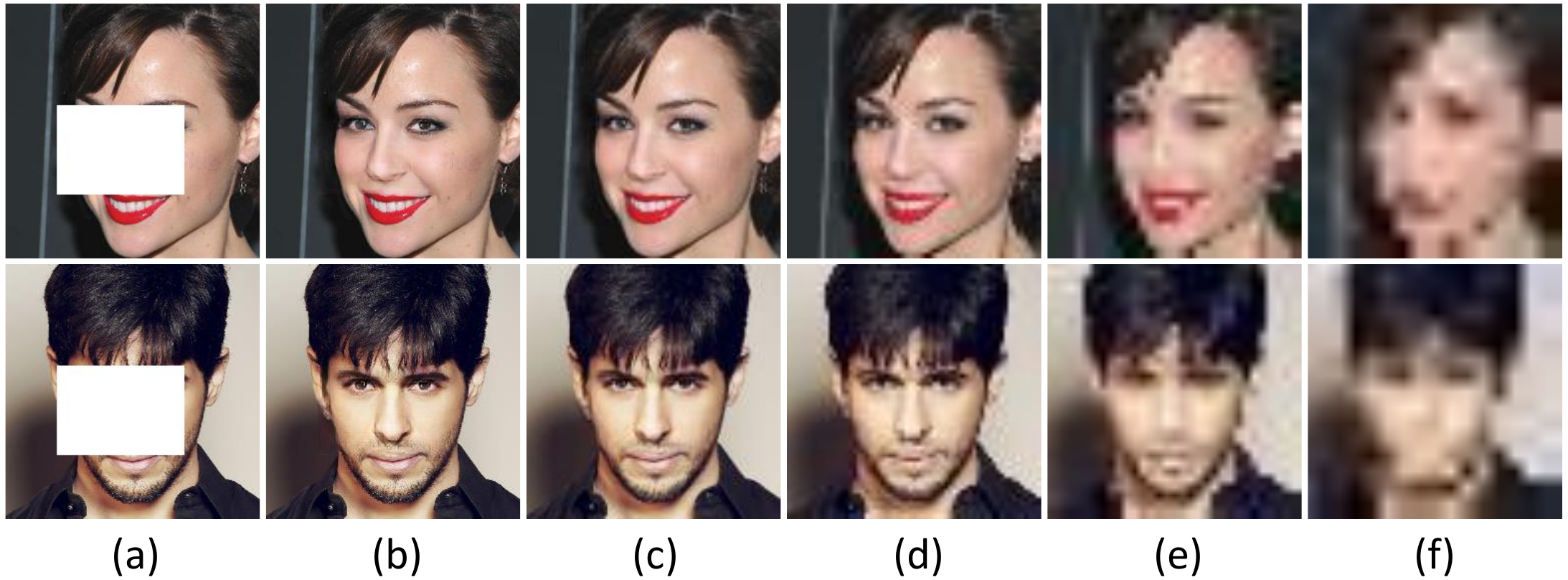}
    \caption{Images produced by the multi-scale decoder. (a) is the input with a rectangle mask. (b) is the final inpainting result. (c), (d), (e) and (f) are outputs at multi-scales.}
    \label{multi}
\end{figure}

The multi-scale decoder can progressively refine the inpainting results at each scale. The experiments are conducted on test images of CelebA-HQ. Then we visualize the images predicted by decoder at several scales (i.e., from $1_{st}$ to $5_{th}$). As shown in Fig. \ref{multi}, it demonstrates that this multi-scale architecture is benefit for decoding learned representations into generated images layer by layer.  

\begin{figure}[ht]
    \centering
    \includegraphics[scale=0.38]{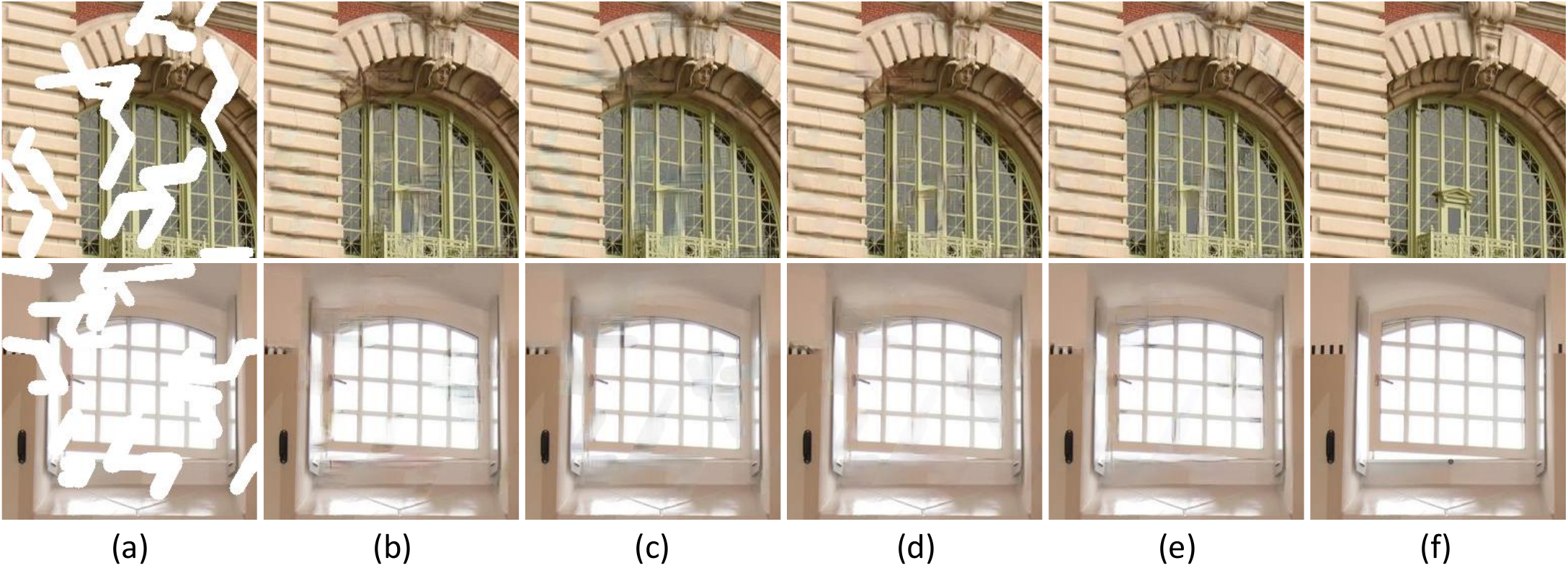}
    \caption{Images produced by the variants of our proposed method. (a) is the input with irregular masks. (e) is the inpainting result generated by the full model. (b), (c) and (d) are generated by the model without DAF and contrastive learning, the model only with contrastive learning and the model only with DAF respectively. (f) is the ground truth.}
    \label{ablation-fig}
\end{figure}

We also investigate the effectiveness of different components of the proposed method. We train several variants of the proposed method: remove the self-supervised Siamese inference network (denote as contrastive learning) and/or the DAF module. As shown in Table \ref{ablation study}, it clearly demonstrates that the inference network and the DAF module play important roles in determining the performance. As shown in Fig. \ref{ablation-fig}, the uncompleted models usually generate obvious artifacts, especially in boundaries while our full model can suppress color discrepancy and artifacts in boundaries and produce realistic inpainting results. 

\section{Conclusion}
In this paper, we propose a novel two-stage paradigm image inpainting method to generate smoother results with reasonable semantics and richer textures. Specifically, the proposed method boosts the ability of the representation learning of the inference network by using contrastive learning. We further design a novel dual attention fusion module to form a multi-scale decoder, which can be embedded into decoder layers in a plug-and-play way. Experiments on CelebA-HQ, Places2, Pairs Street View and ImageNet show the superiority of our proposed method in generating smoother, more coherent and fine-detailed results.

\section{ACKNOWLEDGMENTS}
This work is partially funded by Beijing Natural Science Foundation (Grant No. JQ18017) and Youth Innovation Promotion Association CAS (Grant No. Y201929).

\bibliography{ref}
\bibliographystyle{IEEEtran}




\end{document}